\documentclass[]{spie}  

\def\BibTeX{{\rm B\kern-.05em{\sc i\kern-.025em b}\kern-.08em
    T\kern-.1667em\lower.7ex\hbox{E}\kern-.125emX}}
    
\usepackage{graphicx}
\usepackage{wrapfig} 
\usepackage{amssymb}
\usepackage{amsmath}
\usepackage{amsfonts}
\usepackage{booktabs}
\usepackage{url}
\usepackage{comment}
\usepackage{subcaption}
\usepackage{afterpage}

\everymath{\displaystyle}

\begin{document}

\title{Image quality assessment for determining efficacy and limitations of Super-Resolution Convolutional Neural Network (SRCNN)}
\author{Chris M. Ward, Josh Harguess, Brendan Crabb, Shibin Parameswaran
\\
Space and Naval Warfare Systems Center Pacific 
\\
53560 Hull Street, San Diego, CA 92152-5001
\\
cward@spawar.navy.mil, joshua.harguess@navy.mil, brendancrabb8388@pointloma.edu, shibin.parameswaran@navy.mil  
}

\maketitle
\begin{abstract}
Traditional metrics for evaluating the efficacy of image processing techniques do not lend themselves to understanding the capabilities and limitations of modern image processing methods - particularly those enabled by deep learning. When applying image processing in engineering solutions, a scientist or engineer has a need to justify their design decisions with clear metrics. By applying blind/referenceless image spatial quality (BRISQUE), Structural SIMilarity (SSIM) index scores, and Peak signal-to-noise ratio (PSNR)  to images before and after image processing, we can quantify quality improvements in a meaningful way and determine the lowest recoverable image quality for a given method.
\end{abstract}

\section{Introduction} \label{sec:intro}

Since the inception of digital imagery, there has been a demand for higher and higher resolution images and videos. Image resolution describes the details contained in an image. Higher resolution, in general, means that more information can be extrapolated from the imagery. Two principal areas motivate the desire for higher resolution imagery: improvement of information for human analysis, and performance improvement of machine perception.  While digital image resolution can be classified in different ways: spatial resolution, spectral resolution, temporal resolution, etcetera, this study focuses on methodology for improving spatial resolution.

In recent years, deep learning algorithms and convolutional neural network (CNN) architectures have gained momentum in solving many problems in machine learning (ML) and have, in particular, benefited computer vision research areas. One architecture of interest is the Super-Resolution Convolutional Neural Network (SRCNN), which has demonstrated the application of deep learning to image enhancement.

Additionally, work in no-reference image quality metrics, a growing area of research within computer vision, has produced a relatively new metric model: blind/referenceless image spatial quality evaluator (BRISQUE), which has gained popularity in the evaluation of scene statistics and quantification of loss of “naturalness.” \cite{mittal2012no}

In this paper we will examine SRCNN and its ability to reconstruct imagery degraded by various means. We will quantify our results by applying the BRISQUE metric, and others, and compare the results with qualitative observations. The organization of the paper is as follows: In Section \ref{sec:srcnn} we briefly describe SRCNN and why it was chosen as a vehicle for experimentation in this study. Section \ref{sec:data} and \ref{sec:metrics} introduce the datasets and metrics used for the study, respectively. Section \ref{sec:methodology} describes our methodology, and gives the results of the study. We conclude the paper in Section \ref{sec:conclusion}.

\goodbreak
\section{Experimentation}
\label{sec:experimentation}
In this section, we introduce the datasets used for this paper, briefly explain the methodology for measuring the efficacy of super resolution reconstruction, and present our results.

\subsection{Super-Resolution Convolutional Neural Network (SRCNN)}
\label{sec:srcnn}
SRCNN\cite{dong2014learning} was used as an experimentation vehicle due to its success in recent studies and a growing interest in super-resolution applications. SRCNN was initialized with weights learned from training on the ILSVRC 2013 ImageNet dataset, as described in Dong, Loy, He, and Tong.\cite{dong2014learning}

\subsection{Data}
\label{sec:data}
For this evaluation we utilized familiar imagery from the Set 5 [\cite{bevilacqua2012low}] and Set 14 [\cite{zeyde2010single}] datasets.

\hfill

\begin{figure}[htbp!]
\captionsetup{justification=centering}
\begin{subfigure}{.135\textwidth}
\includegraphics[width=\linewidth]{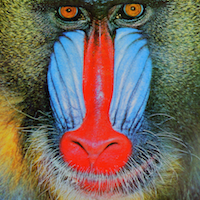}
\caption{set 14\linebreak'baboon'}
\end{subfigure}\hfill
\begin{subfigure}{.135\textwidth}
\includegraphics[width=\linewidth]{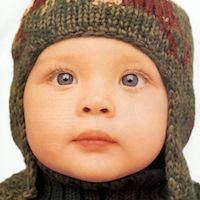}
\caption{set 5\linebreak'baby'}
\end{subfigure}\hfill
\begin{subfigure}{.135\textwidth}
\includegraphics[width=\linewidth]{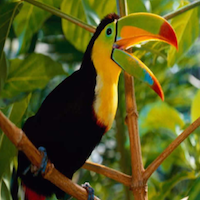}
\caption{set 5\linebreak'bird'}
\end{subfigure}\hfill
\begin{subfigure}{.135\textwidth}
\includegraphics[width=\linewidth]{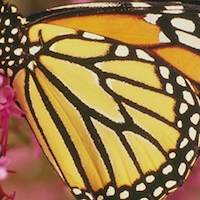}
\caption{set 5\linebreak'butterfly'}
\end{subfigure}\hfill
\begin{subfigure}{.135\textwidth}
\includegraphics[width=\linewidth]{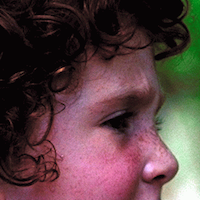}
\caption{set 5\linebreak'head'}
\end{subfigure}\hfill
\begin{subfigure}{.135\textwidth}
\includegraphics[width=\linewidth]{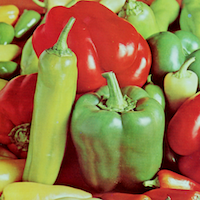}
\caption{set 14\linebreak'pepper'}
\end{subfigure}\hfill
\begin{subfigure}{.135\textwidth}
\includegraphics[width=\linewidth]{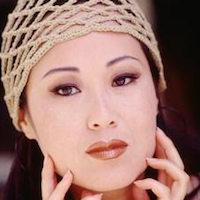}
\caption{set 5\linebreak'woman'}
\end{subfigure}\par

\hfill\linebreak
\caption{Test images sourced from set 5 and set 14}
\label{fig:testImages}
\end{figure}

\subsection{Metrics}
\label{sec:metrics}
The performance of SRCNN in different scenarios was measured  using two full-reference metrics and one no-reference metric. The two full reference metrics used were the peak signal-to-noise ratio (PSNR) and the structural similarity (SSIM) index. A MATLAB implementation of the algorithm for calculating the SSIM index from Wang et al.\cite{wang2004image} was used throughout testing. 

The no-reference metric utilized was the Blind/Referenceless Image Spatial Quality Evaluator(BRISQUE).\cite{mittal2012no}  A MATLAB implementation of this algorithm from Mittal et al. \cite{mittal2011brisque} was used to calculate this metric. PSNR was calculated using the native MATLAB function psnr(A,ref).

\subsubsection{BRISQUE: No-reference Image Quality}
\label{sec:brisque}
BRISQUE is a no-reference metric for evaluating Natural Scene Statistics. No-reference metrics assume that no pristine example of an image is present at the time of assigning a metric its quality.\cite{harguess2017using} However, within the scope of this experiment, we apply this metric to a reference image in order to observe the impact of image reconstruction on the statistical regularity of the imagery under test. Smaller BRISQUE values indicate low distortion and larger values indicate high levels of image distortion. That is to say, the BRISQUE score is inversely proportional to image quality.

BRISQUE was selected as an evaluation metric in this experiment because it was designed for strong correlation to human judgment of quality across different types of distortion.\cite{mittal2012no} Because we aim to evaluate an image-correction method intended to enhance imagery in a subjectively human way, BRISQUE is a particularly appealing metric for this experiment.

Given the commonality of subjective observation between SRCNN and BRISQUE, we hypothesize that we will observe high BRISQUE scores commensurate with imagery degradation, and improved/lower BRISQUE scores for reconstructed imagery.

\subsubsection{SSIM: Structural SIMilarity Index}
\label{sec:ssim}
SSIM is a full-reference metric that describes the statistical similarity of two images. In our experimentation we take a sample image and declare it as a pristine reference, regardless of any preexisting degradation, noise, or anomalies.  Processed imagery is then compared to this reference in the computation of the SSIM  metric.

As described in Wang, Bovik, Sheikh, and Simoncelli \cite{wang2004image}, we use a mean SSIM (MSSIM) index to evaluate the image quality holistically, given that the local application of SSIM provides a more accurate representation of statistical features in an image. Using the SSIM Index, we aim to quantify the similarity between images before and after successive rounds of processing. 

The expected behavior of this metric is a gradually decreasing Index value that is consistent with the amount of applied processing.

\subsubsection{PSNR: Peak Signal-to-Noise Ratio}
\label{sec:psnr}
The Peak Signal-to-Noise Ratio represents the ratio of the reference image pixels to noise pixels introduced during processing. In the scope of this experiment, PSNR is used to evaluate how well processing and correction methods restore a degraded image to its baseline state. Reference images are assumed perfect with zero noise giving them a PSNR score of $\infty$. Processed (reconstructed) images are referenced to their respective baseline images. 

In examination of the reconstructed imagery, a high PSNR indicates an effectively reconstructed image, whereas a low score indicates persistence of distortion after reconstruction.

\subsection{Methodology}
\label{sec:methodology}
For all experiments, SRCNN basic network settings of $f_1=9$, $f_2=1$, $f_3= 5$, $n_1= 64$, and $n_2=32$ were used.\cite{dong2014learning} This  network  was  trained  with  an  upscaling  factor of 3 on over five million sub-images from the ILSVRC 2013 ImageNet detection training partition. To handle color images in the RGB  color space, the network design was adapted to deal with three channels simultaneously by setting the input channels to $c=3$.  This network was trained on the luminance channel in the YCbCr color space as a single-channel $(c=1)$ network; however, previous work\cite{dong2014learning} has showed that training on the Y channel only produced good performance, as measured by average PSNR (dB) scores on color images in the RGB color space because of the high cross-correlation among RGB channels. 

\subsubsection{Image Compression} 
\label{sec:compression}
The  performance  of  SRCNN  on single-image super-resolution was  tested on images with JPEG compression artifacts to gauge its performance at reconstructing poor quality images of different types.  Using a JPEG compression artifact generator, images in our dataset were  given compression artifacts of varying degrees. These images were then  reconstructed using the SRCNN and evaluated using PSNR, SSIM, and BRISQUE.

\subsubsection{Successive Image Correction}
\label{sec:successiveCorrection}
Using the quality metrics PSNR, SSIM, and BRISQUE, the effect of consecutive rounds of SRCNN correction was observed. To begin, each image in our dataset was shrunk and upsampled by a scaling factor of 3 using bicubic interpolation to produce a low-resolution image that was  the same size as the original. This image was then reconstructed using SRCNN and evaluated using our three image quality metrics. Without  rescaling, this image was processed with SRCNN three additional times, observing our key metrics after each consecutive iteration.

\subsubsection{Scaling Factor}
\label{sec:scalingFactor}
The  performance of SRCNN at single-image super resolution was tested on images resized by different scaling factors. Each image in our dataset was shrunk and upsampled by a scaling factor of 2, 3, and 4 using bicubic interpolation to produce a low resolution image that was the same size as the original. This image was then  reconstructed  using  the  SRCNN  and  evaluated  using PSNR, SSIM, and BRISQUE. Using the reconstructed image as the input, this method of resizing and reconstructing was repeated  an  additional  three  times  to  observe  the  effects  of repeated resizing and single-image super resolution.

\subsubsection{Incremental Scaling vs Large Single-Shot Scaling}
\label{sec:incVsSingleshot}

In order to evaluate the ability of SRCNN to correct progressive amounts of interpolation distortion, we processed images scaled by the same factor, but with a varying number of upsampling stages. Once again, we used bicubic interpolation the method of image scaling.

We performed two consective upsampling operations of 2x scaling factor and then evaluated the performance of SRCNN reconstruction on the upsampled imagery. We then upsampled the same images in one operation, but by a scaling factor of 4x. The images were then processed with SRCNN reconstruction, and the results were compared.

\subsection{Results}
This section describes the observations made during our experimentation. In general our results were consistent with our initial assumptions. Each experiment yielded its own interesting data points and anomalies of note.

\newpage
\subsubsection{Effect of Image Compression}
\label{sec:effectofCompression}

\begin{figure}[htbp!]
\captionsetup{justification=centering}

\begin{subfigure}{.30\textwidth}
\includegraphics[width=\linewidth]{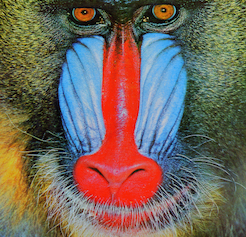}
\caption{Reference Image\linebreak
BRISQUE: 29.9521\linebreak
SSIM: 1\linebreak 
PSNR: $\infty$}
\end{subfigure}\hfill
\begin{subfigure}{.30\textwidth}
\centering
\includegraphics[width=\linewidth]{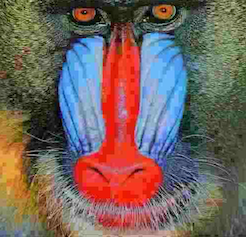}
\caption{Image w/ JPEG Compression\linebreak
BRISQUE: 49.4060\linebreak SSIM: 0.5136\linebreak PSNR: 23.9622}
\label{fig:compressionAddedbaboon}
\end{subfigure}\hfill
\begin{subfigure}{.30\textwidth}
\centering
\includegraphics[width=\linewidth]{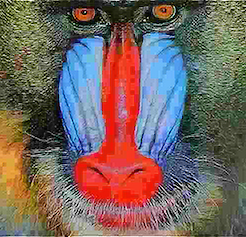}
\caption{SRCNN Reconstruction\linebreak
BRISQUE: 18.1413\linebreak SSIM: 0.3343\linebreak PSNR: 18.3930}
\label{fig:compressionTreatedBaboon}
\end{subfigure}\hfill

\begin{subfigure}{.30\textwidth}
\includegraphics[width=\linewidth]{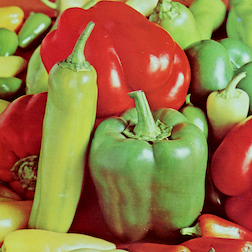}
\caption{Reference Image\linebreak
BRISQUE: 19.3352\linebreak SSIM: 1\linebreak PSNR: $\infty$}
\end{subfigure}\hfill
\begin{subfigure}{.30\textwidth}
\includegraphics[width=\linewidth]{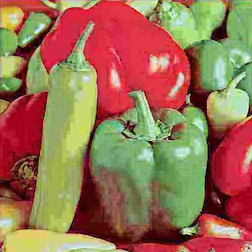}
\caption{Image w/ JPEG Compression\linebreak
BRISQUE: 42.1659\linebreak SSIM: 0.6187\linebreak PSNR: 28.2818}
\label{fig:compressionAddedpepper}
\end{subfigure}\hfill
\begin{subfigure}{.30\textwidth}
\includegraphics[width=\linewidth]{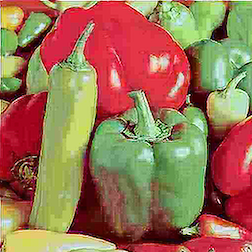}
\caption{SRCNN Reconstruction\linebreak
BRISQUE: 16.9035\linebreak SSIM: 0.4777\linebreak PSNR: 23.3253}
\label{fig:compressionTreatedPepper}
\end{subfigure}\par
\hfill\linebreak
\caption{Effect of compression artifacts}
\label{fig:compressionArtifactsIms}
\end{figure}

\begin{figure}[htbp!]
\captionsetup{justification=centering}
\begin{subfigure}{.99\textwidth}
\includegraphics[width=\linewidth]{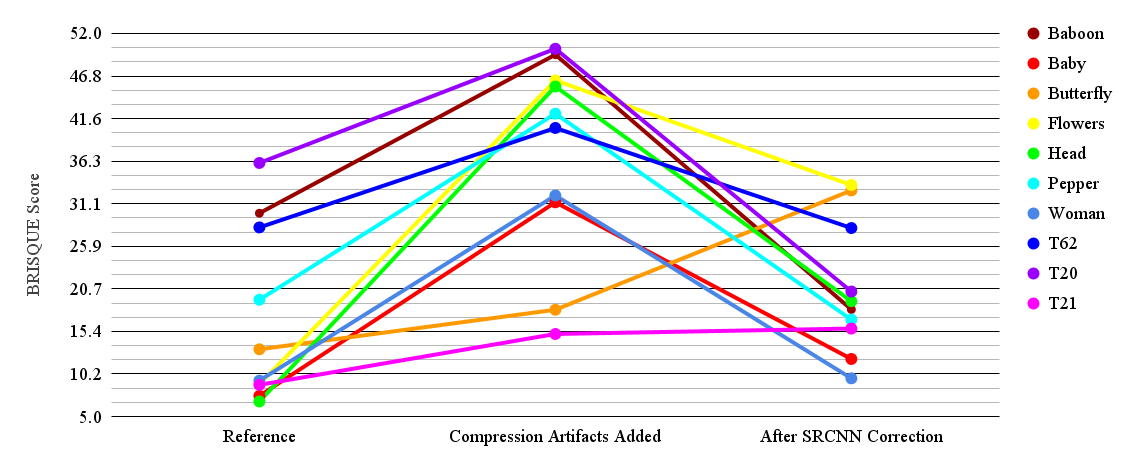}
\caption{BRISQUE}
\label{fig:compressionArtifactsBRISQUE}
\end{subfigure}\hfill

\begin{subfigure}{.99\textwidth}
\includegraphics[width=\linewidth]{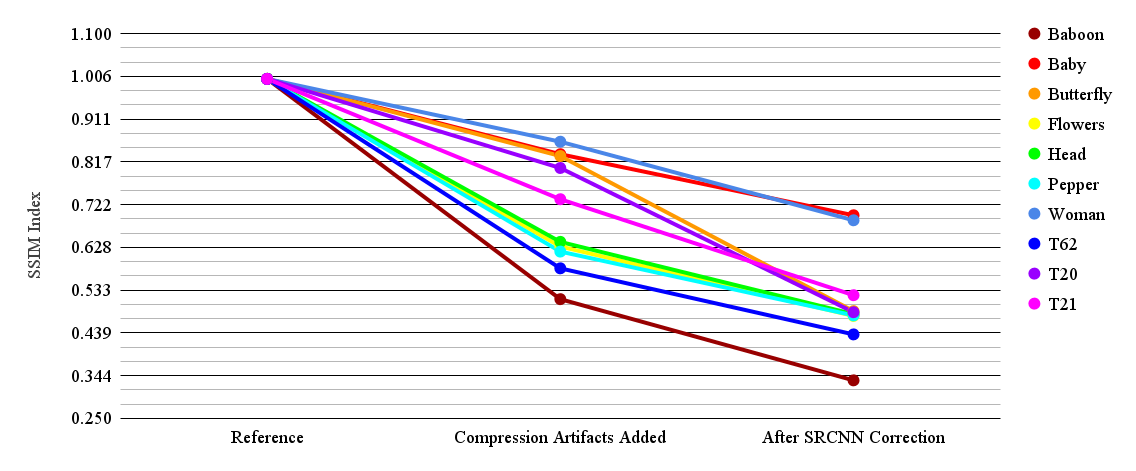}
\caption{SSIM}
\label{fig:compressionArtifactsSSIM}
\end{subfigure}\hfill

\begin{subfigure}{.99\textwidth}
\includegraphics[width=\linewidth]{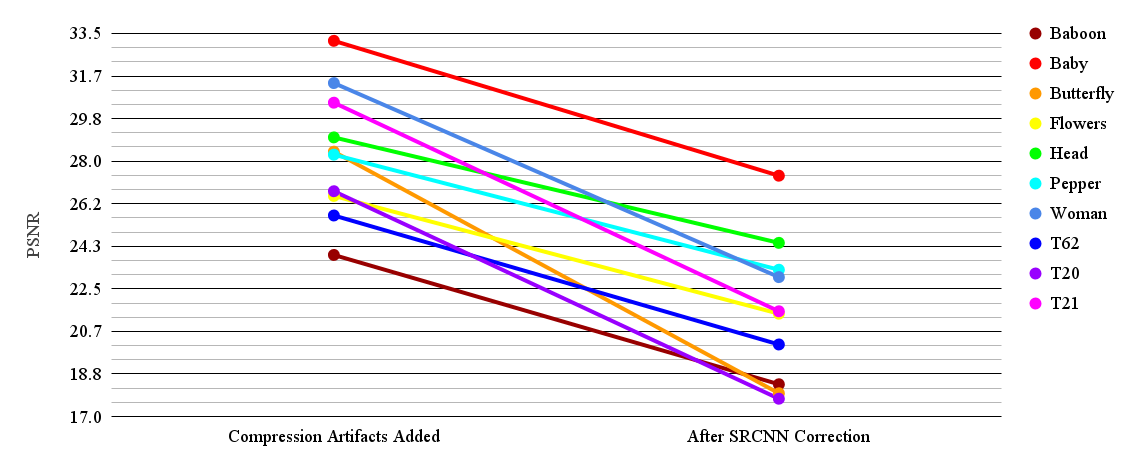}
\caption{PSNR}
\label{fig:compressionArtifactsPSNR}
\end{subfigure}\par

\hfill\linebreak
\caption{Effect of compression artifacts}
\label{fig:compressionArtifactsGrphs}
\end{figure}

The results of applying SRCNN to imagery degraded with JPEG compression artifacts were not as predicted. As expected, insertion of JPEG artifacts degraded each of our metrics - figure \ref{fig:compressionArtifactsSSIM} shows a consistent reduction in structural similarity, and raised levels of noise are apparent in figure \ref{fig:compressionArtifactsPSNR}. Post-interpolation BRISQUE scores also degraded in every image under test [Figure \ref{fig:compressionArtifactsBRISQUE}]. These results are qualitatively observable, noting the visible artifacts present in figures \ref{fig:compressionAddedbaboon} and \ref{fig:compressionAddedpepper}. 

Because SRCNN was not trained to correct compression artifacts, we were uncertain how well it would reconstruct imagery with this tye of degradation. Qualitatively, there was little improvement after reconstruction. Note the persistence of compression artifacts in figures \ref{fig:compressionTreatedBaboon} and \ref{fig:compressionTreatedPepper}. Interestingly, however, SRCNN reconstruction {\bf improved} the BRISQUE scores for nearly every sample tested, besting the score of our reference images in several cases. 

\hfill\linebreak

\clearpage

\subsubsection{Effect of Successive Image Correction}
\begin{figure}[htbp!]
\captionsetup{justification=centering}
\begin{subfigure}{.99\textwidth}
\includegraphics[width=\linewidth]{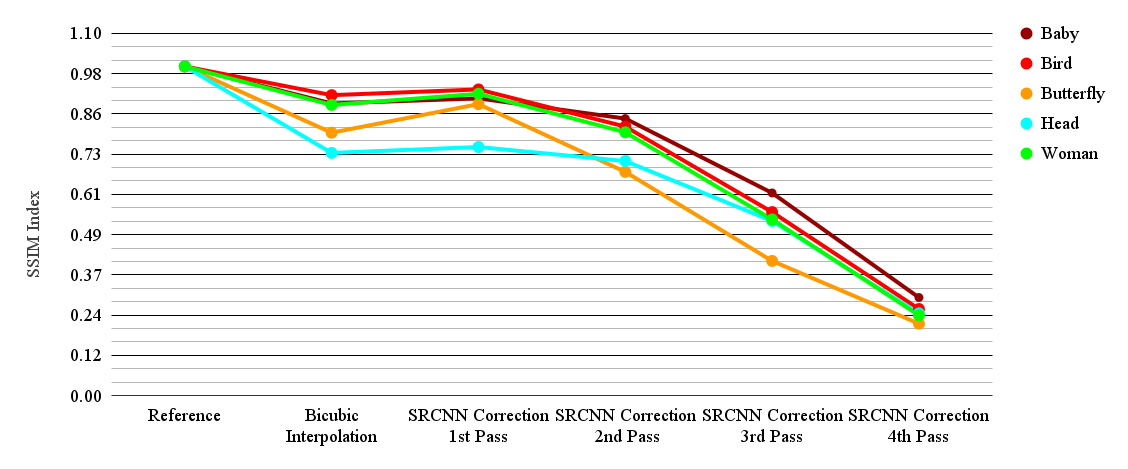}
\caption{1x Bicubic Interpolation Factor}
\end{subfigure}\hfill

\begin{subfigure}{.99\textwidth}
\includegraphics[width=\linewidth]{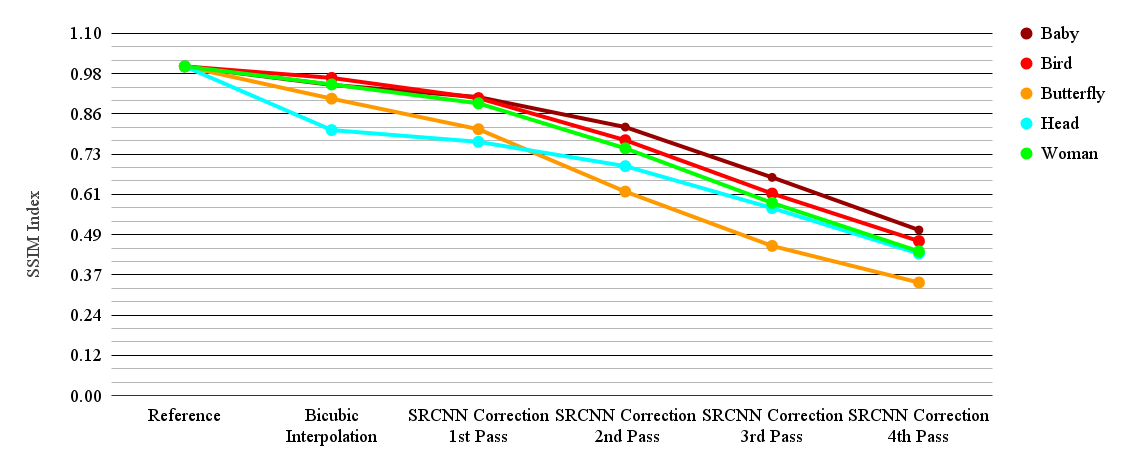}
\caption{2x Bicubic Interpolation Factor}
\end{subfigure}\hfill

\begin{subfigure}{.99\textwidth}
\includegraphics[width=\linewidth]{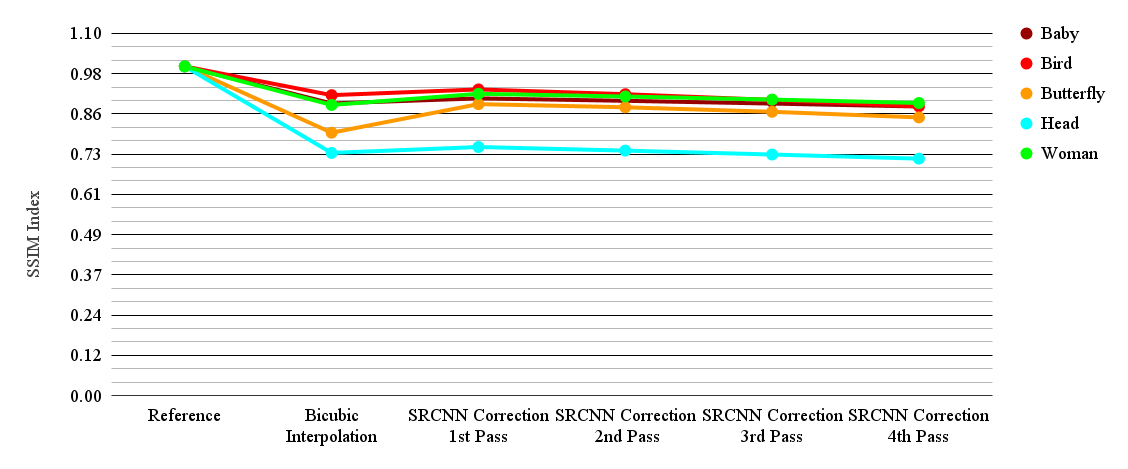}
\caption{3x Bicubic Interpolation Factor}
\end{subfigure}\par

\hfill\linebreak
\caption{Effects of SRCNN on SSIM Index}
\label{fig:SSIMresults}
\end{figure}

\begin{figure}[htbp!]
\captionsetup{justification=centering}
\begin{subfigure}{.99\textwidth}
\includegraphics[width=\linewidth]{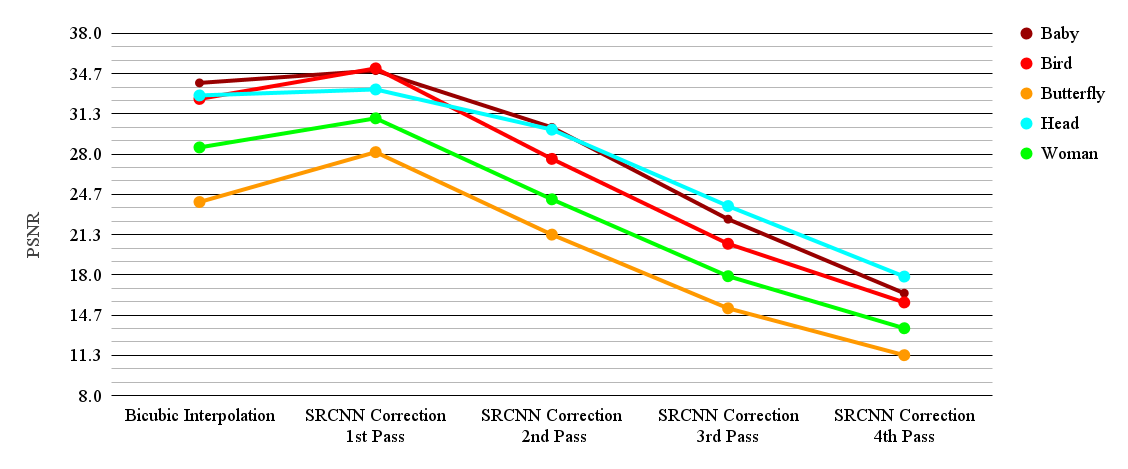}
\caption{1x Bicubic Interpolation Factor}
\end{subfigure}\hfill

\begin{subfigure}{.99\textwidth}
\includegraphics[width=\linewidth]{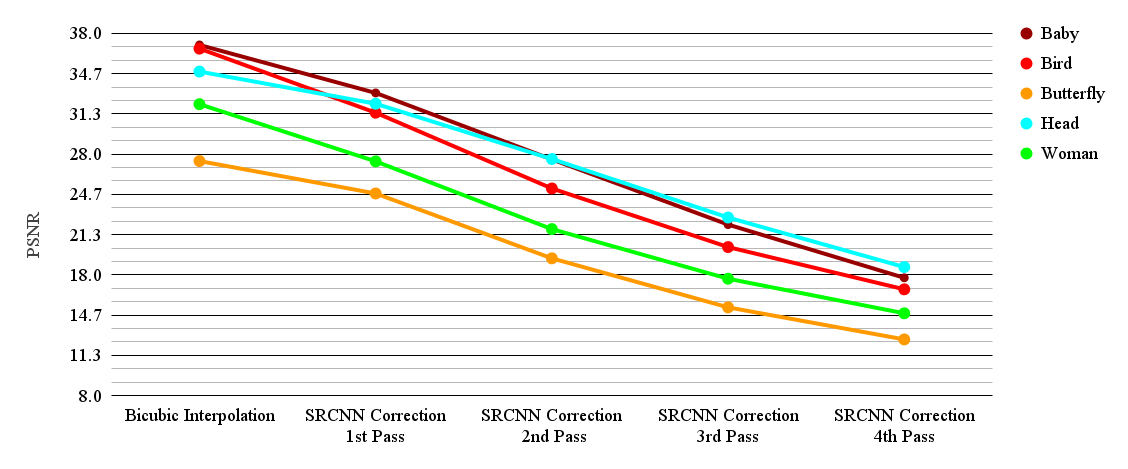}
\caption{2x Bicubic Interpolation Factor}
\end{subfigure}\hfill

\begin{subfigure}{.99\textwidth}
\includegraphics[width=\linewidth]{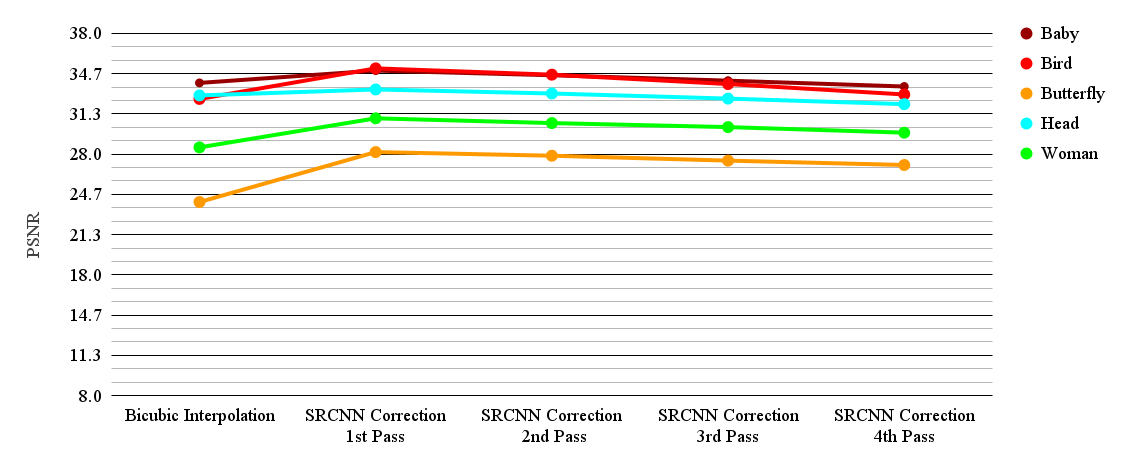}
\caption{3x Bicubic Interpolation Factor}
\end{subfigure}\par

\hfill\linebreak
\caption{Effects of SRCNN on PSNR}
\label{fig:PSNRresults}
\end{figure}

Both BRISQUE and SSIM metrics were degraded by interpolation and, as expected, SRCNN reconstruction improved BRISQUE and PSNR measurements. While each image tested saw improved BRISQUE scores after a singles reconstruction pass, subsequent passes yielded nominal improvement and ultimately had a negative impact (Figure \ref{fig:PSNRresults}). Single-pass reconstruction had no significant impact on structure similarity, but as seen in figure \ref{fig:SSIMresults} there was significant structure change with each subsequent iteration. In section \ref{sec:effectScalingFactor} we discuss how scaling factor influences these trends.

\begin{figure}[htbp!]
\captionsetup{justification=centering}
\begin{subfigure}{.162\textwidth}
\includegraphics[width=\linewidth]{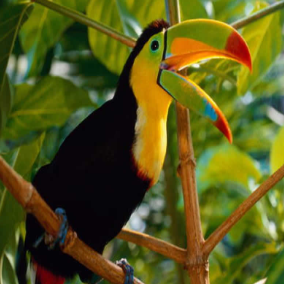}
\caption{\scriptsize Reference Image\linebreak
BRISQUE: 29.7588\linebreak SSIM: 1\linebreak PSNR: $\infty$ }
\end{subfigure}\hfill
\begin{subfigure}{.162\textwidth}
\includegraphics[width=\linewidth]{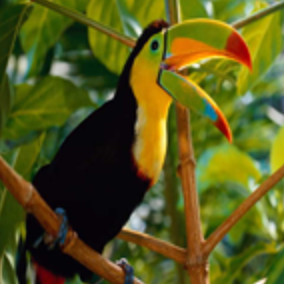}
\caption{\scriptsize Interpolated\linebreak
 BRISQUE: 43.5097\linebreak SSIM: 0.9652\linebreak PSNR: 36.7943}
\end{subfigure}\hfill
\begin{subfigure}{.162\textwidth}
\includegraphics[width=\linewidth]{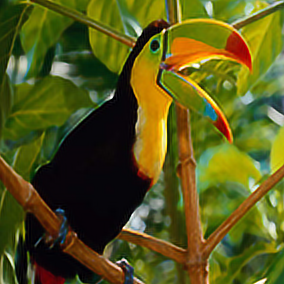}
\caption{\scriptsize SRCNN(1\textsuperscript{st} Pass)\linebreak
BRISQUE: 18.2822\linebreak SSIM: 0.9037\linebreak PSNR: 31.4394}
\end{subfigure}\hfill
\begin{subfigure}{.162\textwidth}
\includegraphics[width=\linewidth]{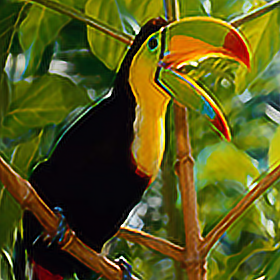}
\caption{\scriptsize SRCNN(2\textsuperscript{nd} Pass)\linebreak
BRISQUE: 19.0244\linebreak SSIM: 0.7759\linebreak PSNR: 25.1624}
\end{subfigure}\hfill
\begin{subfigure}{.162\textwidth}
\includegraphics[width=\linewidth]{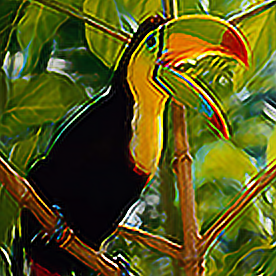}
\caption{\scriptsize SRCNN(3\textsuperscript{rd} Pass)\linebreak
BRISQUE: 30.2409\linebreak SSIM: 0.6138\linebreak PSNR: 20.3081}
\end{subfigure}\hfill
\begin{subfigure}{.162\textwidth}
\includegraphics[width=\linewidth]{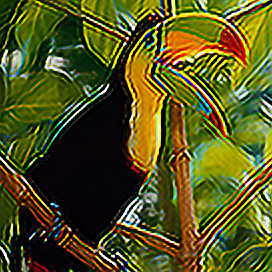}
\caption{\scriptsize SRCNN(4\textsuperscript{th} Pass)\linebreak
BRISQUE: 48.0252\linebreak SSIM: 0.4694\linebreak PSNR: 16.8065}
\label{fig:successiveResults_ringingBird}
\end{subfigure}\hfill

\begin{subfigure}{.162\textwidth}
\includegraphics[width=\linewidth]{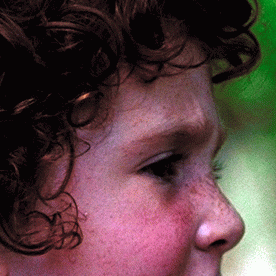}
\caption{\scriptsize Reference Image\linebreak
BRISQUE: 6.8600\linebreak SSIM: 1\linebreak PSNR: $\infty$ }
\end{subfigure}\hfill
\begin{subfigure}{.162\textwidth}
\includegraphics[width=\linewidth]{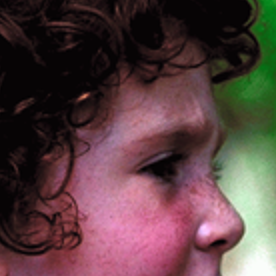}
\caption{\scriptsize Interpolated\linebreak
 BRISQUE: 28.5007\linebreak SSIM: 0.80676\linebreak PSNR: 34.8459}
\end{subfigure}\hfill
\begin{subfigure}{.162\textwidth}
\includegraphics[width=\linewidth]{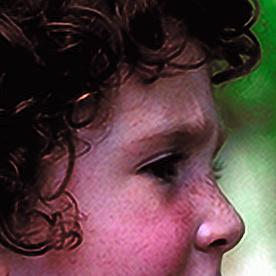}
\caption{\scriptsize SRCNN(1\textsuperscript{st} Pass)\linebreak
BRISQUE: 21.4105\linebreak SSIM: 0.7709\linebreak PSNR: 32.1834}
\end{subfigure}\hfill
\begin{subfigure}{.162\textwidth}
\includegraphics[width=\linewidth]{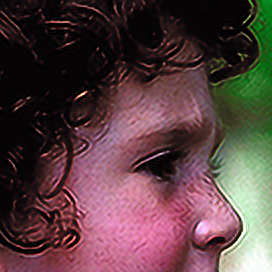}
\caption{\scriptsize SRCNN(2\textsuperscript{nd} Pass)\linebreak
BRISQUE: 29.8259\linebreak SSIM: 0.6971\linebreak PSNR: 27.6023}
\end{subfigure}\hfill
\begin{subfigure}{.162\textwidth}
\includegraphics[width=\linewidth]{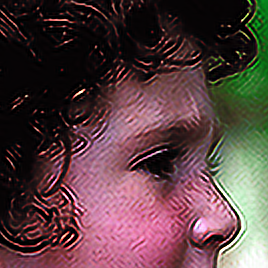}
\caption{\scriptsize SRCNN(3\textsuperscript{rd} Pass)\linebreak
BRISQUE: 36.8977\linebreak SSIM: 0.5697\linebreak PSNR: 22.7547}
\end{subfigure}\hfill
\begin{subfigure}{.162\textwidth}
\includegraphics[width=\linewidth]{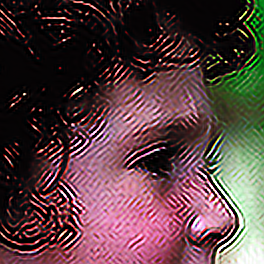}
\caption{\scriptsize SRCNN(4\textsuperscript{th} Pass)\linebreak
BRISQUE: 43.0000\linebreak SSIM: 0.4312\linebreak PSNR: 18.6518}
\label{fig:successiveResults_ringingHead}
\end{subfigure}\hfill

\begin{subfigure}{.162\textwidth}
\includegraphics[width=\linewidth]{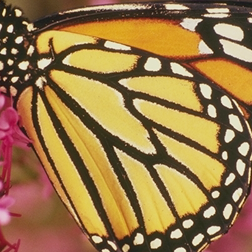}
\caption{\scriptsize Reference Image\linebreak
BRISQUE: 13.2584\linebreak SSIM: 1\linebreak PSNR: $\infty$ }
\end{subfigure}\hfill
\begin{subfigure}{.162\textwidth}
\includegraphics[width=\linewidth]{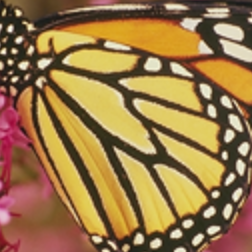}
\caption{\scriptsize Interpolated\linebreak
BRISQUE: 34.1258\linebreak SSIM: 0.9022\linebreak PSNR: 27.4325}
\end{subfigure}\hfill
\begin{subfigure}{.162\textwidth}
\includegraphics[width=\linewidth]{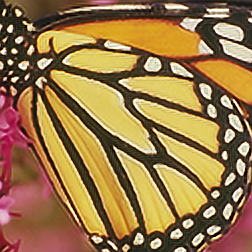}
\caption{\scriptsize SRCNN(1\textsuperscript{st} Pass)\linebreak
BRISQUE: 28.2522\linebreak SSIM: 0.8093\linebreak PSNR: 24.7472}
\end{subfigure}\hfill
\begin{subfigure}{.162\textwidth}
\includegraphics[width=\linewidth]{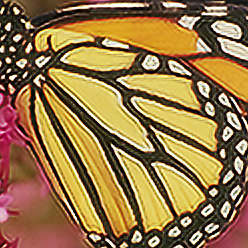}
\caption{\scriptsize SRCNN(2\textsuperscript{nd} Pass)\linebreak
BRISQUE: 62.2542\linebreak SSIM: 0.6197\linebreak PSNR: 19.3736}
\end{subfigure}\hfill
\begin{subfigure}{.162\textwidth}
\includegraphics[width=\linewidth]{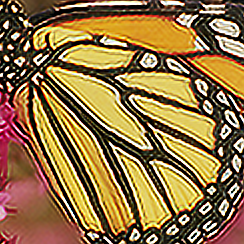}
\caption{\scriptsize SRCNN(3\textsuperscript{rd}Pass)\linebreak
BRISQUE: 95.2858\linebreak SSIM: 0.4547\linebreak PSNR: 15.3125}
\end{subfigure}\hfill
\begin{subfigure}{.162\textwidth}
\includegraphics[width=\linewidth]{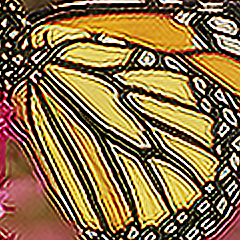}
\caption{\scriptsize SRCNN(4\textsuperscript{th} Pass)\linebreak
BRISQUE: 140.783\linebreak SSIM: 0.3435\linebreak PSNR: 12.6557}
\label{fig:successiveResults_ringingButterfly}
\end{subfigure}\par

\hfill\linebreak
\caption{Effect of successive SRCNN correction at 2x Interpolation factor}
\label{fig:successiveResults}
\end{figure}

Qualitative observations show that 'sharpness' increases with each reconstruction pass. At 2x interpolation factor we begin to see 'ringing' about edges after three passes through SRCNN. This edge-ringing can be clearly observed after four passes in figures \ref{fig:successiveResults_ringingBird}, \ref{fig:successiveResults_ringingHead}, and \ref{fig:successiveResults_ringingButterfly}. Based on the Gibbs phenomena, we attribute this effect to the progressive approximation of a discontinuous function by SRCNN. Increasing BRISQUE scores confirm a loss of 'naturalness' in our test imagery.

\hfill\linebreak

\newpage

\begin{figure}[htbp!]
\captionsetup{justification=centering}
\begin{subfigure}{.245\textwidth}
\includegraphics[width=\linewidth]{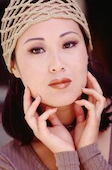}
\caption{\scriptsize Reference Image\linebreak
BRISQUE: 9.4141\linebreak SSIM: 1\linebreak PSNR: $\infty$ }
\end{subfigure}\hfill
\begin{subfigure}{.245\textwidth}
\includegraphics[width=\linewidth]{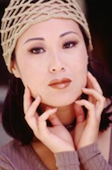}
\caption{\scriptsize 1/2 Scale \linebreak
BRISQUE: 33.1319\linebreak SSIM: 0.9454\linebreak PSNR: 32.1447}
\end{subfigure}\hfill
\begin{subfigure}{.245\textwidth}
\includegraphics[width=\linewidth]{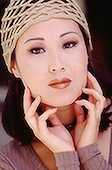}
\caption{\scriptsize 2x Scale w/SRCNN(1\textsuperscript{st} Pass)\linebreak
BRISQUE: 7.6792\linebreak SSIM: 0.8880\linebreak PSNR: 27.4067}
\end{subfigure}\hfill
\begin{subfigure}{.245\textwidth}
\includegraphics[width=\linewidth]{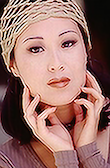}
\caption{\scriptsize 2x Scale w/SRCNN(2\textsuperscript{nd} Pass)\linebreak
BRISQUE: 12.1168\linebreak SSIM: 0.7513\linebreak PSNR: 21.7961}
\end{subfigure}\hfill
\begin{subfigure}{.245\textwidth}
\includegraphics[width=\linewidth]{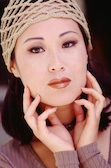}
\caption{\scriptsize Reference Image\linebreak
BRISQUE: 9.4141\linebreak SSIM: 1\linebreak PSNR: $\infty$ }
\end{subfigure}\hfill
\begin{subfigure}{.245\textwidth}
\includegraphics[width=\linewidth]{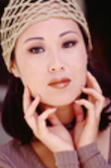}
\caption{\scriptsize 1/3 Scale \linebreak
BRISQUE: 52.2679\linebreak SSIM: 0.8830\linebreak PSNR: 28.5632}
\end{subfigure}\hfill
\begin{subfigure}{.245\textwidth}
\includegraphics[width=\linewidth]{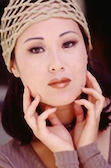}v
\caption{\scriptsize 3x Scale w/SRCNN(1\textsuperscript{st} Pass)\linebreak
BRISQUE: 32.4458\linebreak SSIM: 0.9169\linebreak PSNR: 30.9741}
\end{subfigure}\hfill
\begin{subfigure}{.245\textwidth}
\includegraphics[width=\linewidth]{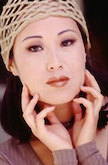}v
\caption{\scriptsize 3x Scale w/SRCNN(2\textsuperscript{nd} Pass)\linebreak
BRISQUE: 30.9157\linebreak SSIM: 0.9086\linebreak PSNR: 30.5790}
\end{subfigure}\hfill

\begin{subfigure}{.245\textwidth}
\includegraphics[width=\linewidth]{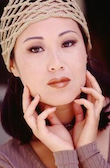}
\caption{\scriptsize Reference Image\linebreak
BRISQUE: 9.4141\linebreak SSIM: 1\linebreak PSNR: $\infty$ }
\end{subfigure}\hfill
\begin{subfigure}{.245\textwidth}
\includegraphics[width=\linewidth]{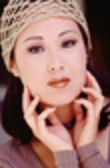}
\caption{\scriptsize 1/4 Scale \linebreak
 BRISQUE: 57.345\linebreak SSIM: 0.8210\linebreak PSNR: 26.4652}
\end{subfigure}\hfill
\begin{subfigure}{.245\textwidth}
\includegraphics[width=\linewidth]{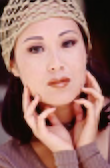}
\caption{\scriptsize 4x Scale w/SRCNN(1\textsuperscript{st} Pass)\linebreak
BRISQUE: 50.9325\linebreak SSIM: 0.8424\linebreak PSNR: 27.3121}
\end{subfigure}\hfill
\begin{subfigure}{.245\textwidth}
\includegraphics[width=\linewidth]{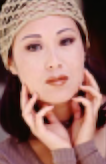}
\caption{\scriptsize 4x Scale w/SRCNN(2\textsuperscript{nd} Pass)\linebreak
BRISQUE: 55.0843\linebreak SSIM: 0.8260\linebreak PSNR: 26.7147}
\end{subfigure}\par

\hfill\linebreak
\caption{Effect of Scaling Factor}
\label{fig:effectScaling}
\end{figure}


\begin{figure}[htbp!]
\captionsetup{justification=centering}
\begin{subfigure}{.99\textwidth}
\includegraphics[width=\linewidth]{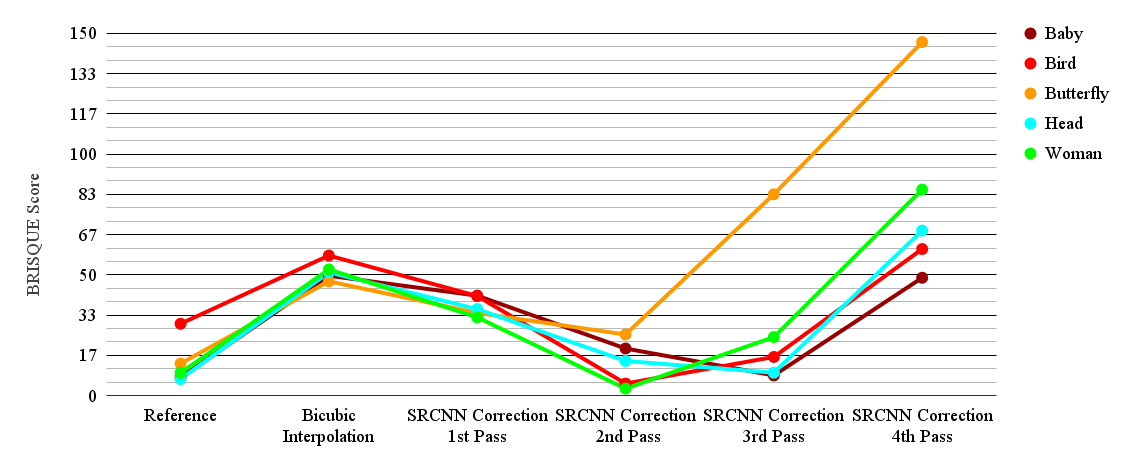}
\caption{1x Bicubic Interpolation Factor}
\end{subfigure}\hfill

\begin{subfigure}{.99\textwidth}
\includegraphics[width=\linewidth]{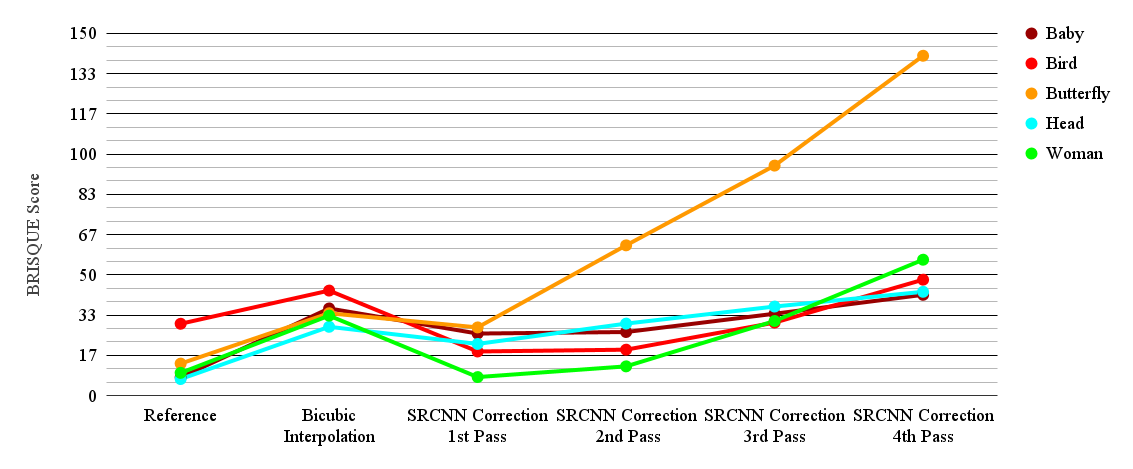}
\caption{2x Bicubic Interpolation Factor}
\end{subfigure}\hfill

\begin{subfigure}{.99\textwidth}
\includegraphics[width=\linewidth]{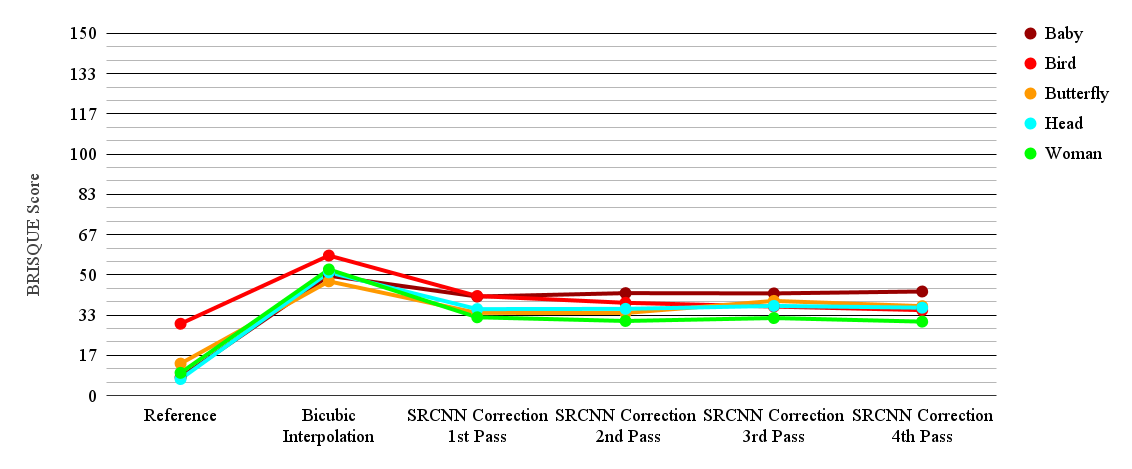}
\caption{3x Bicubic Interpolation Factor}
\end{subfigure}\par

\hfill\linebreak
\caption{Effects of SRCNN on BRISQUE Score}
\label{fig:BRISQUEresults}
\end{figure}

\newpage

\newpage

\clearpage

\begin{figure}[htbp!]
\captionsetup{justification=centering}
\begin{subfigure}{.247\textwidth}
\includegraphics[width=\linewidth]{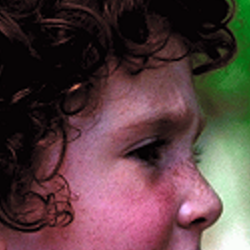}
\caption{Reference Image
\linebreak BRISQUE: 6.0740}
\end{subfigure}\hfill
\begin{subfigure}{.247\textwidth}
\includegraphics[width=\linewidth]{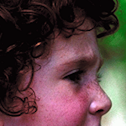}
\caption{4x Bicubic Downsample\linebreak BRISQUE: 56.3197}
\end{subfigure}\hfill
\begin{subfigure}{.247\textwidth}
\includegraphics[width=\linewidth]{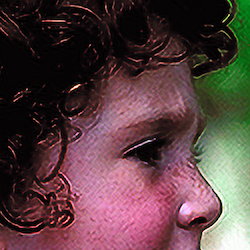}
\caption{2x2x Scale w/SRCNN\linebreak
BRISQUE: 35.1189}
\end{subfigure}\hfill
\begin{subfigure}{.247\textwidth}
\includegraphics[width=\linewidth]{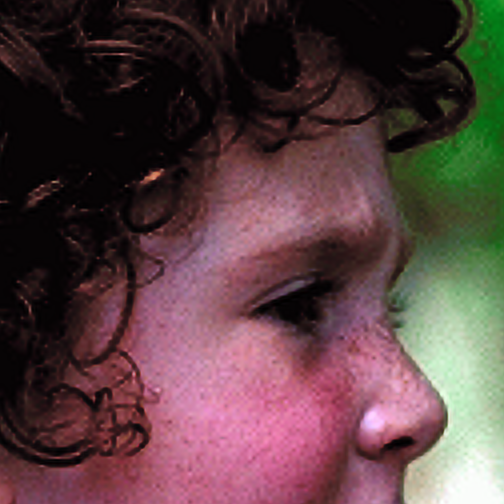}
\caption{4x Scale w/SRCNN\linebreak BRISQUE: 60.6498}
\end{subfigure}\hfill

\begin{subfigure}{.247\textwidth}
\includegraphics[width=\linewidth]{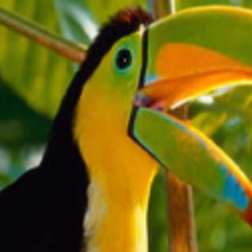}
\caption{Reference Image
\linebreak BRISQUE: 16.9799}
\label{fig:birdOG}
\end{subfigure}\hfill
\begin{subfigure}{.247\textwidth}
\includegraphics[width=\linewidth]{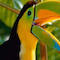}
\caption{4x Bicubic Downsample\linebreak BRISQUE: 70.4471}
\end{subfigure}\hfill
\begin{subfigure}{.247\textwidth}
\includegraphics[width=\linewidth]{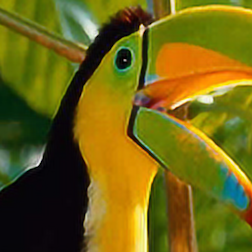}
\caption{2x2x Scale w/SRCNN\linebreak
BRISQUE: 16.607}
\label{fig:better1}
\end{subfigure}\hfill
\begin{subfigure}{.247\textwidth}
\includegraphics[width=\linewidth]{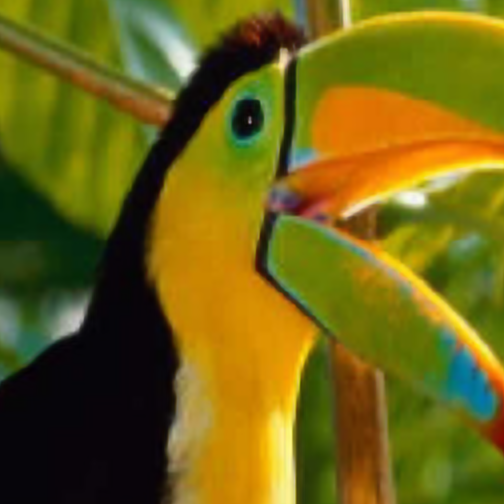}
\caption{4x Scale w/SRCNN\linebreak BRISQUE: 63.1386}
\end{subfigure}\hfill

\begin{subfigure}{.247\textwidth}
\includegraphics[width=\linewidth]{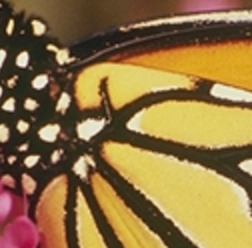}
\caption{Reference Image
\linebreak BRISQUE: 21.4839}
\label{fig:butterflyOG}
\end{subfigure}\hfill
\begin{subfigure}{.247\textwidth}
\includegraphics[width=\linewidth]{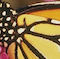}
\caption{4x Bicubic Downsample\linebreak BRISQUE: 59.7064}
\end{subfigure}\hfill
\begin{subfigure}{.247\textwidth}
\includegraphics[width=\linewidth]{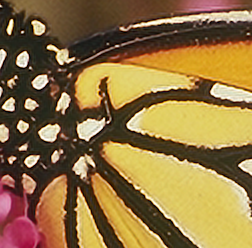}
\caption{2x2x Scale w/SRCNN\linebreak
BRISQUE: 17.9103 }
\label{fig:better2}
\end{subfigure}\hfill
\begin{subfigure}{.247\textwidth}
\includegraphics[width=\linewidth]{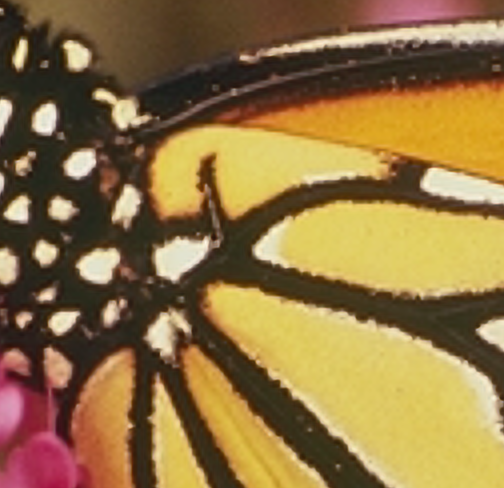}
\caption{4x Scale w/SRCNN\linebreak BRISQUE: 49.9278}
\end{subfigure}\hfill

\begin{subfigure}{.247\textwidth}
\includegraphics[width=\linewidth]{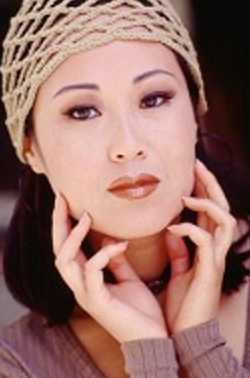}
\caption{Reference Image\linebreak BRISQUE: 12.9570}
\end{subfigure}\hfill
\begin{subfigure}{.247\textwidth}
\includegraphics[width=\linewidth]{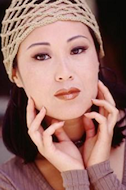}
\caption{4x Downsample\linebreak BRISQUE: 55.4144}
\end{subfigure}\hfill
\begin{subfigure}{.247\textwidth}
\includegraphics[width=\linewidth]{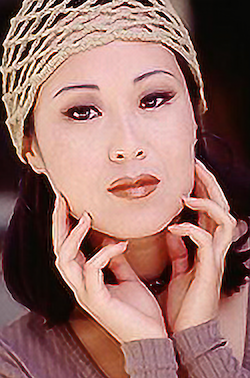}
\caption{2x2x Scale w/SRCNN
\linebreak BRISQUE: 23.3831}
\end{subfigure}\hfill
\begin{subfigure}{.247\textwidth}
\includegraphics[width=\linewidth]{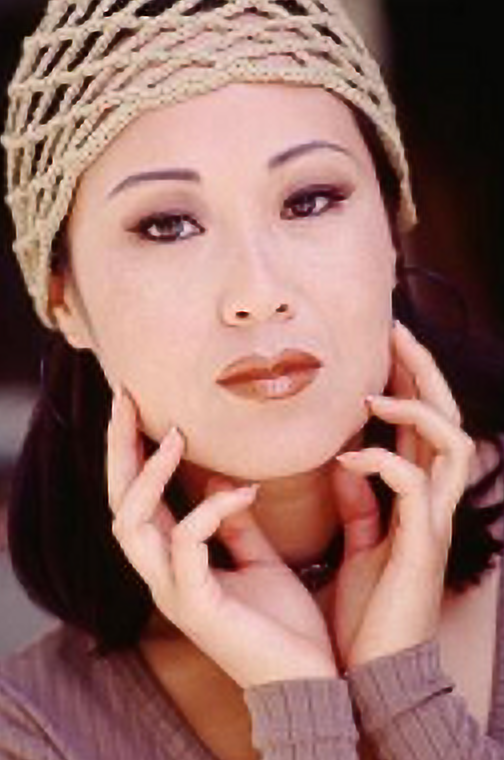}
\caption{4x Scale w/SRCNN\linebreak BRISQUE: 51.0999}
\end{subfigure}\par
\hfill\linebreak
\caption{Efficacy of correction on incremental upsampling vs large single-shot upsampling}
\label{fig:incrementalVsingleshot}
\end{figure}


\clearpage

\subsubsection{Effect of Scaling Factor} 
\label{sec:effectScalingFactor}
Examination of data in Figures \ref{fig:BRISQUEresults}, \ref{fig:SSIMresults}, and \ref{fig:PSNRresults} show the effect of scaling factor on our test imagery. As plotted, we note that increased scaling factor mitigates the efficacy of SRCNN and it effects on BRISQUE, SSIM, and PSNR. This is particularly true in the 3rd and 4th reconstruction passes. Moreover, this observation holds true qualitatively. In figure \ref{fig:effectScaling} we see that images scaled by a factor of two, have a much more visual response to SRCNN. When compared to imagery scaled by a 3x factor, we see that the sharpening/enhancement is much more prominent in the 2x test case. Imagery scaled at 4x is even less responsive to SRCNN reconstruction passes.

\subsubsection{Efficacy of Reconstruction on Incremental Upsampling vs Large Single-Shot Upsampling}
The efficacy of SRCNN reconstruction on incrementally scaled imagery (2x2x) outperformed 4x single-shot scaling in every test case. Examination of Figure \ref{fig:incrementalVsingleshot} shows much sharper details present in all 2x2x cases. Quantitatively, the BRISQUE measurements of 2x2x test cases agree with visual inspection, yielding better scores than 4x in each tested image. In several particularly noteworthy cases (Figures \ref{fig:better1} and \ref{fig:better2}), SRCNN reconstruction yielded BRISQUE scores that surpass those of their corresponding reference images (Figures \ref{fig:birdOG} and \ref{fig:butterflyOG}).

\section{Conclusion and Future Work}
\label{sec:conclusion}
In this paper we have applied three different metrics (BRISQUE, SSIM, and PSNR) to imagery that has been modified by varying means, and then reconstructed using the Super-Resolution Convolutional Neural Network (SRCNN). SRCNN reconstructed images as expected, sharpening imagery affected by bicubic interpolation. The approach of using the BRISQUE algorithm to evaluate SRCNN revealed that SRCNN successfully restores the 'naturalness' of imagery, but is not without limitation. Additionally, we observed the role that image scaling factor plays on the efficacy of SRCNN. 

We also observed that other types of distortion, such as JPEG compression artifacts, are not only resistant to SRCNN reconstruction, but produce erratic BRISQUE scores. One area of future work is to study the 'gaussianess' of these images to better understand why the BRISQUE metric improved despite the persistence of compression artifacts after SRCNN reconstruction.

We are interested in using BRISQUE and SRCNN to further study future image processing neural networks, and advance work in adversarial imagery \cite{harguess2017using} by attempting to correct adversarial features with SRCNN. In general, we hope to work toward building more robust metrics for analyzing deep-learning architectures for image processing, and use them in the development of high performing deep learning architectures.

\bibliography{iqa-metrics-SPIESD18}   
\bibliographystyle{spiebib}   

\end{document}